%% file: Template.tex
\documentclass{article}
\usepackage{spconf,amsmath,graphicx}
\usepackage{multirow}
\usepackage{amssymb}
\usepackage{enumitem}
\usepackage{hyperref}
\usepackage{algorithm}
\usepackage[noend]{algpseudocode}

\title{Gradient-Aware Logit Adjustment Loss for Long-tailed Classifier}
\name{Fan Zhang$^1$ $^6$, Wei Qin$^2$, Weijieying Ren$^3$, Lei Wang$^4$, Zetong Chen$^5$, Richang Hong$^2$ }
  
\address{
    $^1$ AHU-IAI AI Joint Laboratory, Anhui University \\
    $^2$ Hefei University of Technology \\
    $^3$ Pennsylvania State University \\
    $^4$ Singapore Managerment University \\
    $^5$ University of Science and Technology of China \\
    $^6$ Institute of Artificial Intelligence, Hefei Comprehensive National Science Center
}
\begin{document}
\maketitle
\input{sections/0_abs}

\input{sections/1_intro}
\input{sections/3_method}
\input{sections/4_exp}

\input{sections/5_conclusion}

\bibliographystyle{IEEEbib}
\bibliography{strings,refs}

\end{document}

%% file: sections/0_abs.tex
\begin{abstract}
In the real-world setting, data often follows a long-tailed distribution, where head classes contain significantly more training samples than tail classes. Consequently, models trained on such data tend to be biased toward head classes. The medium of this bias is imbalanced gradients, which include not only the ratio of scale between positive and negative gradients but also imbalanced gradients from different negative classes. 
Therefore, we propose the Gradient-Aware Logit Adjustment (GALA) loss, which adjusts the logits based on accumulated gradients to balance the optimization process. Additionally, We find that most of the solutions to long-tailed problems are still biased towards head classes in the end, and we propose a simple and post hoc prediction re-balancing strategy to further mitigate the basis toward head class.
Extensive experiments are conducted on multiple popular long-tailed recognition benchmark datasets to evaluate the effectiveness of these two designs. Our approach achieves top-1 accuracy of 48.5\%, 41.4\%, and 73.3\% on CIFAR100-LT, Places-LT, and iNaturalist, outperforming the state-of-the-art method GCL by a significant margin of 3.62\%, 0.76\% and 1.2\%, respectively. Code is available at \href{https://github.com/lt-project-repository/lt-project}{https://github.com/lt-project-repository/lt-project}.
\end{abstract}


%% file: sections/1_intro.tex
\section{Introduction}
\label{intro}

Deep learning methods have achieved promising performance in various domains \cite{qin_image} due to the availability of large-scale data sources.
However, deploying deep learning models to real-world scenarios usually achieves unsatisfactory results since realistic data mainly exhibit a long-tailed distribution, where head classes dominate the majority of samples and tail classes contain only a few samples \cite{menon2021longtail, Iscen2021ClassBalancedDF}.
The invisibility of tail class samples makes the model learn insufficient information and thus is biased toward head classes~\cite{Wang2022TowardsCH, Kang2020Decoupling}. 

The problem of long-tailed data has been a growing concern within the deep learning community in recent years. In response, Kang et al.~\cite{Kang2020Decoupling} find that adjusting classifiers may be more important for long-tail problems. Compared with label frequency~\cite{{ren2020balanced}}, prediction~\cite{Lin2017FocalLF}, and some learnable weight~\cite{zhang2021distribution}, adjusting gradients is a more straightforward way to influence classifier. This paper investigates how imbalanced gradients bias the classifier toward head classes as shown in Fig.~\ref{fig:problem}(a).

As a preliminary, we first assume the classifier is a feed-forward network where each class has its learnable weight vector (i.e., class vector). We then introduce two distinct gradients for each class vector: positive gradients, obtained from samples of this class, pull this class vector close to samples' features within this class, and negative gradients, obtained from samples of other classes, push this class vector away from samples' features of other classes.
\input{figures/problem}
Imbalanced gradients distort the classifier in two ways: \textbf{(i) 
The imbalanced ratio of positive and negative gradients (called gradient ratio in short) for tail class vectors causes the classifier to identify samples belonging to tail classes as other classes.}
As shown by the blue curve in Fig.~\ref{fig:problem}(b), since a tail class contains fewer samples, for its class vector positive gradients are much smaller than negative gradients.
Most of the negative gradients come from head classes, thus the tail class vector will be pushed away from the samples from these head classes. As for samples from other classes, they are very close together in the feature space.
\textbf{(ii) For each class vector, negative gradients from different classes are imbalanced. It is the undiscovered devil resulting in the class vector misclassifying other tail class samples as its class.}
For a class vector, the negative gradients from tail class samples are much smaller than the negative gradients from head class samples as shown in Fig.~\ref{fig:problem}(c). The smaller negative gradients from other tail class samples cannot push the class vector away enough from these samples.
Thus, the imbalanced negative gradients for a particular class vector result in that other tail class samples are misclassified as its class.

To mitigate the classifier's bias caused by imbalanced gradients, we propose a novel loss function called Gradient-Aware Logit Adjustment (GALA) loss, by introducing two margin items into the linear classifier of each class.
These two margin items collaborate to balance gradients for class vectors. One item is to adjust the total negative gradients to ensure an appropriate gradient ratio. Another item seeks to balance the negative gradients from different classes. With the two margin items, as shown in Fig.~\ref{fig:problem}(a), Fig.~\ref{fig:problem}(b), Fig.~\ref{fig:problem}(c), and Fig.~\ref{fig:problem}(d) respectively, the norm of classifier weights is more balanced, the ratio of positive and negative gradients is more balanced, the negative gradients from different classes are more balanced, and the similarity between a tail class vector and the samples from its class is larger.

Although our proposed loss function is effective in reducing biases in classifier, there may still be biases in predictions towards head classes due to other components, such as the CNN. To address this issue, we utilize a simple yet effective test-time strategy called prediction re-balance strategy. This strategy can directly mitigate prediction biases, regardless of whether they originate from biased classifiers or biased CNNs. All of these biased components ultimately contribute to prediction bias. Our re-balance strategy normalizes predictions, making the prediction gap across classes less disparate.

Our approach achieves $48.5\%$, $55.0\%$, $41.4\%$, and $73.3\%$ top-1 accuracy on CIFAR100-LT (IF=100), ImageNet-LT, Places-LT, and iNaturalist, which surpasses the SOTA method, i.e., GCL~\cite{li_gcl} by an obvious margin of $3.62\%$, $0.12\%$, $0.76\%$ and $1.2\%$, respectively.
In summary, the key contributions of our work are three-fold:
\begin{itemize}
    \item We uncover how imbalanced accumulated gradients result in biased classifiers. Based on our observation, we propose a novel Gradient-Aware Logit Adjustment loss function, which uses accumulated gradients to regulate the gradients of the classifier weights.
    \item We propose a simple but effective and general prediction re-balance strategy to mitigate the remaining prediction biases.
    \item By theoretical analysis and empirical experiments, we show that our proposed GALA loss could balance the gradient. Experiments performed on several datasets and ablation studies show the superior performance and effectiveness of the proposed GALA loss and the prediction re-balanced strategy.
\end{itemize}

%% file: figures/problem.tex
\begin{figure*}[t]
\begin{center}
\includegraphics[width=1.0\textwidth]{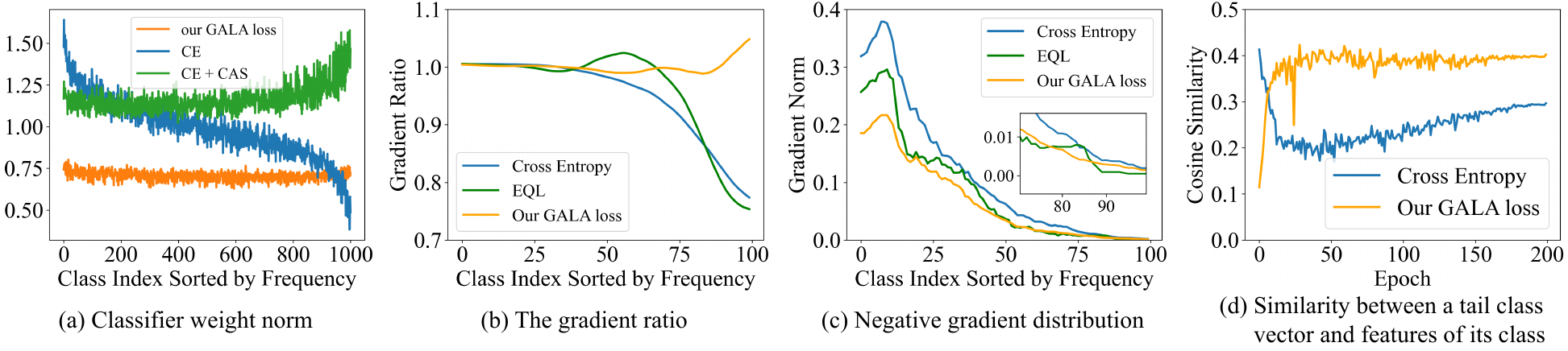}
\end{center}
\vspace{-9mm}
\caption{(a) demonstrates the imbalanced weight norms of long-tailed biased classifiers. The full name of CAS~\cite{Kang2020Decoupling} is class aware sampler which is used to balance the number of samples in each class. Statistics are from experiments performed on ImageNet-LT. (b) reports the gradient ratio (the ratio of positive to negative gradients) of Cross Entropy, EQL~\cite{Tan2020EqualizationLF}, and our GALA loss. (c) reports the imbalanced negative gradients from different classes of Cross Entropy, EQL, and our GALA loss. (d) describes the average similarity between a tail class vector and features of its class when the model is trained with Cross Entropy and our GALA loss. Statistics for (b), (c), and (d) are from experiments performed on CIFAR100-LT.}
\label{fig:problem}
\vspace{-5mm}
\end{figure*}

%% file: sections/3_method.tex
\section{Method}
\label{sec_method}

\noindent\textbf{Problem Setup and Notations.}
In long-tailed $K$-way classification tasks, we train models on a imbalanced training set $\mathcal{D}=\{ (x_i, y_i)\}_{i=1}^{N}$, where $y_i \in [1,..., K]$ is the annotated label of sample $x_i$.
$D_k$ denotes the set of all training examples of $k$-th class and $|D_k|$ is the cardinality of the class. The imbalance factor, IF$=\frac{max_k |D_k|}{min_k |D_k|}$, evaluates how imbalanced the dataset is. For simplicity, we use $x$ to denote the feature of a sample from the feature extractor (CNN) in this paper. The classifier of the $j$-th class is defined as:
\begin{align}
\small  
    \mathcal{F}^{(j)}(x) &= \omega_{j}^{T}{x}+b_j
\end{align}
where $\omega_j$ and $b_j$ present the weight (the class vector we mentioned before) and the bias term of the $j$-th classifier. The $\mathcal{F}^{j}(x)$ denotes the logit over the $j$-th class. The probability over the $j$-th is computed by Softmax:
\begin{align}
\small
    p^{(j)}(x) &= \frac{\exp{(\mathcal{F}^{(j)}(x)})}{\sum_{l=1}^{K} \exp{(\mathcal{F}^{(l)}(x))}}
\end{align}
where the sum of the probabilities for each class over all classes should be 1. We measure the prediction error by cross-entropy (CE) loss. Since the ground-truth label is the $k$-th class, the loss is formulated as:
\begin{align}
\small
    \mathcal{L}_{ce} &= - \log p^{(k)}(x) \\
    &= - \log \frac{\exp{(\mathcal{F}^{(k)}(x)})}{\sum_{l=1}^{K} \exp{(\mathcal{F}^{(l)}(x))}} \label{eq_log_softmax}
\end{align}
Based on the loss function $\mathcal{L}_{ce}$, we obtain the gradient of the class vector $\omega_{j}$ and optimize $\omega_j$ by an optimizer (e.g., SGD optimizer). The optimization could be formulated as:
\begin{align}
\small
    & \omega_{j} = \omega_j - lr * \sum_{x \in \mathcal{X}} \frac{\partial \mathcal{L}_{ce}(x)}{\partial \omega_{j}}
\end{align}where $\mathcal{X}$ denotes the training samples and $lr$ denotes the learning rate.
\subsection{Gradient-Aware Logit Adjustment Loss Function}
\noindent \textbf{The proposed GALA loss.} We add two gradient-based terms into the linear classifier to adjust the logits for each class (to simplify the equations, the bias term $b_j$ is ignored.):
\begin{align}
\small
    \mathcal{F}^{(j)}(x)  =  
    \begin{cases}
            \omega_{j}^{T}x + \log \theta_{j} - \log \phi_{k}, \quad &j \neq k \\
            \omega_{j}^{T}x, \quad &j = k 
    \end{cases}            
      \label{eq_bal_fc}
\end{align}
where $\mathcal{F}^{(j)}$ is the linear classifier for the class $j$ and $\omega_{j}$ is the class weight of $\mathcal{F}^{j} $. $x$ denotes the feature of a sample and we assume its annotated label is the $k$-th class.
$\theta_j$ is the accumulated positive gradients for the class weight $\omega_j$ during training. 
With introduced {term} $\theta_j$, negative gradients for the weight $\omega_j$ will be positively associated with its accumulated positive gradient. Thus, this {term} could balance the positive gradient and the negative gradient of each class weight.
$\phi_{k}$ is the accumulated negative gradients produced from samples of the $k$-th class.
If $\phi_{k}$ is too large, it would suppress the negative gradients from the $k$-th class samples in the later training and vice versa.
Thus, this term could balance the gradient from different negative classes. Notably, all gradients are obtained by accumulating the gradients generated by the logits of each epoch.

\noindent\textbf{Theoretical Analysis.} We will verify that our GALA loss could balance both the gradient ratio and the negative distributions by theoretical analysis. Based on the eq.~\ref{eq_log_softmax} and eq.~\ref{eq_bal_fc}, the GALA loss function could be reformulated as:
\begin{align}
    \mathcal{L}_{\text{GALA}} =& - \log \frac{\exp{(\mathcal{F}^{(k)}(x)})}{\sum_{l=1}^{K} \exp{(\mathcal{F}^{(l)}(x))}} \\
    =& \log(1 + \sum_{l \neq k}\exp{(\mathcal{F}^{l}(x)-\mathcal{F}^{k}(x))}) \\
    =& \log(1 + \sum_{l \neq k}\exp{(\omega_{l}^{T}x + \log \theta_{l} - \log \phi_{k}-\mathcal{F}^{k}(x))})  \\
    =& \log(1 + \exp{(\omega_{l^{'}}^{T}x + \log \theta_{l^{'}} - \log \phi_{k}-\mathcal{F}^{k}(x))} \nonumber \\ 
    +& \sum_{(l \neq k) \& (l \neq l^{'})}\exp{(\omega_{l}^{T}x + \log \theta_{l} - \log \phi_{k}-\mathcal{F}^{k}(x))})  
    \label{eq_final_loss}
\end{align}
where $x$ is the feature of a $k$-th class sample.
Next we calculate the negative gradient for $\omega_{l^{'}}$,  i.e., the weight of $l^{'}$-th classifier, from the $k$-th class sample($l^{'} \neq k$). To simplify the derivation process, we regard some unrelated items in eq.~\ref{eq_final_loss} as a constant $\alpha$:
\begin{align}
    \mathcal{L}_{\text{GALA}} = \log(\exp{(\omega_{l^{'}}x+\log \theta_{l^{'}} - \log \phi_k - \mathcal{F}^{k}(x)))}+\alpha) 
\end{align}
Next we calculate the negative gradient for $\omega_{l^{'}}$, i.e., the class weight of $l^{'}$-th classifier, from the $k$-th class sample ($l^{'} \neq k$).
\begin{align}
    \frac{\partial \mathcal{L}_{\text{GALA}}(x)}{\partial\omega_{l^{'}}} =&  \frac{1}{\exp{(\omega_{l^{'}}x+\log \theta_{l^{'}} - \log \phi_k- \mathcal{F}^{k}(x))}+\alpha}*  \nonumber  \\
    & \frac{\partial \, ({\exp{(\omega_{l^{'}}x+\log \theta_{l^{'}} - \log \phi_k- \mathcal{F}^{k}(x)))}+\alpha})}{\partial{\omega_{l^{'}}}}  \nonumber \\
    =& \frac{\exp{(\omega_{l^{'}}x+\log \theta_{l^{'}} - \log \phi_k- \mathcal{F}^{k}(x)))}}{\exp{(\omega_{l^{'}}x+\log \theta_{l^{'}} - \log \phi_k- \mathcal{F}^{k}(x)))}+\alpha} x  \nonumber \\
    =& \frac{x}{1+\frac{\alpha * \phi_k}{\exp{(\omega_{l^{'}}x  - \mathcal{F}^{k}(x)))}*\theta_j}} \label{final_gradient}
\end{align}
From eq.~\ref{final_gradient}, we can find that, if the model is trained with our GALA loss, the negative gradient of the $l^{'}$-th class weight $\frac{\partial\mathcal{L}}{\partial\omega_{l^{'}}}$ is positively associated with $\theta_{j}$ and negatively associated with $\phi_{k}$.
Thus, we can conclude that our GALA loss with these two terms could balance both the gradient ratio and the negative gradient from different classes.

\subsection{Prediction Re-balancing Strategy}
\noindent \textbf{Motivation.}
\input{figures/figure_pos_pred_method} Due to imbalanced data, the predictions are biased toward head classes. As shown in Fig.~\ref{fig:pos_pred_method}, when the model is naively trained on CIFAR100-LT (IF=100), more samples are predicted as head classes on a balanced testset. 
Therefore, we could re-balance the prediction probabilities to mitigate the prediction bias.

\noindent \textbf{The prediction re-balance.}
Formally, let the prediction probability matrix $\mathbf{P} = \{p_{bk}\} \in \mathcal{R}^{B \times K }  $, where $B$ is the size of the test set and $K$ is the number of the classes. $p_{*k} \in \mathcal{R}^{B}$ are the probabilities that all test samples are predicted to be the $k$-th class. We scale the prediction probabilities of $\mathbf{P}$ to get $\widetilde{\mathbf{P}} = \{\widetilde{p}_{*k}\}$ by:
\begin{align}
    \widetilde{p}_{*k} = \frac{p_{*k}}{\lVert p_{*k} \rVert^{\tau}}
\end{align}
where $\tau$ is a hyper-parameter controlling the "temperature" of the normalisation~\cite{Hinton2015DistillingTK}, and $\lVert \cdot \rVert$ denotes the $L_1$ norm. When $\tau = 1$, it reduces to standard $L_1$-normalization. When $\tau=0$, no normalization is imposed. After the prediction probabilities normalization, we choose the class with the highest normalized probability for each sample as the prediction for this sample.


%% file: figures/figure_pos_pred_method.tex
\begin{figure}[t]
\begin{center}
\includegraphics[width=0.3\textwidth]{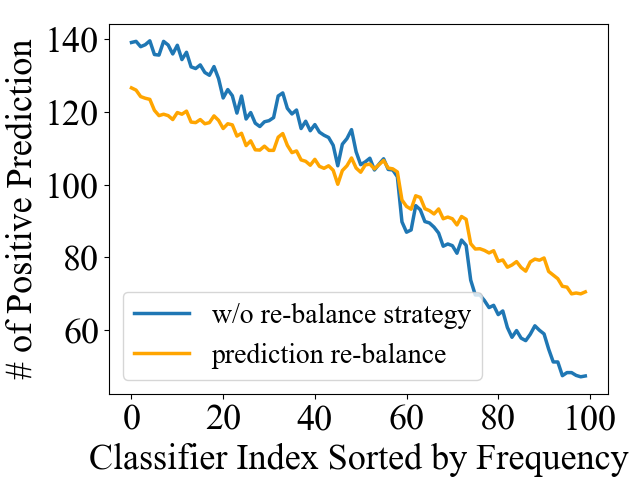}
\end{center}
\vspace{-8mm}
\caption{The number of positive predictions (the number of samples is predicted as its
class over all classes) of different methods.}
\label{fig:pos_pred_method}
\vspace{-5mm}
\end{figure}

%% file: sections/4_exp.tex
\section{Experiments}
\label{sec_exp}
\subsection{Setup}
We perform experiments on several long-tailed image classification datasets, including CIFAR100-LT, ImageNet-LT, Places-LT~\cite{liu2019large} and iNaturalist2018~\cite{van2018inaturalist}. Datasets with the "LT" suffix are long-tailed versions of their original datasets and iNaturalist 2018 is a real-world fine-grained classification dataset.
For all experiments, we use the SGD optimizer with momentum 0.9 and cosine learning rate scheduler to optimize networks. Besides, we train each model for 200 epochs, with a batch size of 64 (for CIFAR100-LT and ImageNet-LT)/256(for iNaturalist). All settings not mentioned are the same as MisLAS~\cite{zhong2021improving}.
\input{tables/tab_cifar100.tex}
\subsection{Main results}
\label{sec_sub_results}
Comparative experiments have been performed to show the superior performance of our GALA loss. The comparison results on CIFAR100-LT, ImageNet-LT, Places-LT, and iNaturalist are presented in Table~\ref{tab_cifar}, Table~\ref{tab_three_dataset}.
Some results are produced by the official codes and some results are quoted from related papers. 

\noindent \textbf{Results on CIFAR100-LT.}
We conducted extensive experiments on the CIFAR100-LT dataset with four different imbalance factors: 200, 100, 50, and 10. The comparison results are summarized in Table~\ref{tab_cifar}. Our proposed GALA loss outperforms many prior methods by obvious margins with all imbalance factors, especially for the severe imbalanced factors. In specific, our proposed approach achieves $48.5\%$, $52.3\%$, and $56.5\%$ in top-1 classification accuracy for imbalance factor 200, 100, and 50, which surpass the second best method, i.e., GCL by a significant margin of $3.62\%$, $3.59\%$ and $2.95\%$, respectively.

\input{tables/tab_three_dataset}
\noindent \textbf{Results on large-scale datasets.}
Experiment results performed on ImageNet-LT, iNaturalist2018, and Places-LT are reported in Table~\ref{tab_three_dataset}. {Under the fair training setting (the same network architecture and training epochs), our proposed GALA loss achieves the best performance, which outperforms a recent SOTA work GCL~\cite{li_gcl}}.

%% file: tables/tab_cifar100.tex
\begin{table}[t]
\caption{Comparison results on CIFAR100-LT in terms of top-1 accuracy($\%$). The backbone architecture is ResNet-32.}
\vspace{1mm}
\label{tab_cifar}{%
\begin{tabular}{ccccc}

\hline
Imbalance Factor(IF)       & 200 & 100 & 50 & 10 \\ \hline
CE loss                & 34.84  & 38.43  & 43.9  & 55.71 \\
LDAM-DRW~\cite{cao_ldam}         & 38.91  & 42.04  & 47.62 & 58.71 \\
TDE~\cite{tang_TDE}  & -      & 44.1   & 50.3  & 59.6  \\
mixup + cRT~\cite{Kang2020Decoupling}      & 41.73  & 45.12  & 50.86 &       \\
BBN~\cite{zhou2020bbn}              & 37.21  & 42.56  & 47.02 & 59.12 \\
Equalization loss~\cite{Tan2020EqualizationLF}&    43.38    &   -     &  -     & -\\
Balanced Softmax~\cite{ren2020balanced} & 43.3   & 45     & 50.5  & 61.1     \\
MisLAS~\cite{zhong2021improving}          & 42.33  & 47.5   & 52.62 & 63.2  \\
DiVE~\cite{He2021DistillingVE}             &    -   & 45.35  & 51.13 & 62    \\
GCL~\cite{li_gcl}              & 44.88  & 48.71  & 53.55 &   -   \\
Weight Balancing~\cite{alshammari2022long}   & 40.8   & 44.3   & 49.8  &    -  \\ \hline
GALA loss \tiny(ours) & 48.0   & 52.1   & 56.3  & 64.0  \\
GALA loss + PN & \textbf{{48.5}}   & \textbf{{52.3}}   & \textbf{{56.5}}  & \textbf{{64.2}}  \\
\hline
\end{tabular}%
\vspace{-4mm}
}
\end{table}

%% file: tables/tab_three_dataset.tex
\begin{table}[t]
\caption{Comparison results on ImageNet-LT, iNaturalist, and Places-LT in terms of top-1 accuracy($\%$).}
\vspace{1mm}
\label{tab_three_dataset}
{%
\setlength{\tabcolsep}{0.6mm}{
\begin{tabular}{cccc}
\hline
           & ImageNet-LT & iNaturalist & \multicolumn{1}{l}{Places-LT} \\ \hline
baseline         & 46.8        & 61.7        & 30.2                          \\
LDAM-DRW~\cite{cao_ldam}         & 51.2        & 68          & -                             \\
cRT~\cite{Kang2020Decoupling}              & 50.3        & 65.2        & 36.7                          \\
Balanced Softmax~\cite{ren2020balanced} & 47.2        & -           & 38.7                          \\
TDE~\cite{tang_TDE}              & 51.3        & 68.7        & -                             \\
DRO-LT~\cite{samuel2021distributional}           & 53.5        & 69.7        & -                             \\
DisAlign~\cite{Xu2022ConstructingBF}         & 53.4        & 70.6        & 39.3                          \\
Logit adjustment~\cite{menon2021longtail} & 51.11       & 71.98       & -                             \\
Weight Balancing~\cite{alshammari2022long} & 53.9        & 70.2        & -                             \\
DLSA~\cite{Xu2022ConstructingBF}             & -           & 71.6        & 38.2                          \\
MiSLAS~\cite{zhong2021improving}            & 53.4        & 71.6        & 40.4                          \\
GCL~\cite{li_gcl}              & 54.8        & 72.1        & 40.64                         \\
vMF~\cite{Wang2022TowardsCH}              & 53.4        & 71          & -                             \\ \hline
GALA loss \tiny(ours)        & 53.4       & 71.2        & 41.0  \\
GALA loss + PN \tiny(ours)        & \textbf{{55.0}}        & \textbf{{73.3}}        & \textbf{{41.4}}                          \\ \hline
\end{tabular}
}
\vspace{-4mm}
}
\end{table}

%% file: sections/5_conclusion.tex
\section{Conclusion}
\label{sec_conclu}
In this paper, we analyze how imbalanced gradients, including both imbalanced 
positive/negative gradient ratio and imbalanced gradients from different negative classes, undermine the optimization of classifiers and result in a biased model. 
Based on our analysis, a balanced gradient margin (GALA) loss is designed which is theoretically shown to be able to balance these two types of imbalanced gradients simultaneously. 
To further mitigate prediction bias toward head classes, we introduce a post hoc technique named prediction re-balancing strategy, which directly normalizes the prediction probabilities over classes. Extensive experiments on various benchmark datasets have demonstrated the effectiveness of these two designs, and our method achieves superior performances compared with other methods.